\documentclass[11pt]{article}
\usepackage{acl2015}
\usepackage{times}
\usepackage{url}
\usepackage{latexsym}
\usepackage{booktabs}
\usepackage{multirow}
\usepackage{amsmath} 
\usepackage{hyperref}
\title{SemEval-2024 Task 8: Weighted Layer Averaging RoBERTa for Black-Box Machine-Generated Text Detection}

\author{Ayan Datta$^\dagger$ \\
  IIIT Hyderabad \\
  {\tt ayan.datta}\\{\tt @research.iiit.ac.in} \\\And
  Aryan Chandramania$^\dagger$ \\
  IIIT Hyderabad \\
  {\tt aryan.chandramania}\\{\tt @research.iiit.ac.in} \\\AND
  Radhika Mamidi \\
  IIIT Hyderabad \\
  {\tt radhika.mamidi@iiit.ac.in}}

\date{}

\begin{document}
\maketitle
\begin{abstract}
 This document contains the details of the authors' submission to the proceedings of SemEval 2024's Task 8: Multigenerator, Multidomain, and Multilingual Black-Box Machine-Generated Text Detection Subtask A (monolingual) and B. Detection of machine-generated text is becoming an increasingly important task, with the advent of large language models (LLMs). In this paper, we lay out how using weighted averages of RoBERTa layers lets us capture information about text that is relevant to machine-generated text detection.
\end{abstract}
\def\thefootnote{$\dagger$}\footnotetext{These authors contributed equally to this work.}


\section{Introduction}

Language modeling is a foundational task in NLP, and encompasses learning of all the features that make up language. Different levels of linguistic information are stored in language models' (LM) hidden states. This may include syntax, morphological features, phrasing, and so on. \cite{10.1162/tacl_a_00349} Our aim is to leverage this encoded information to help us discern machine-generated text. 

The advent of large language models (LLMs) has transformed the digital landscape, and this has also led to the proliferation of machine-generated text in spaces spanning from legal proceedings, to articles, to school submissions. With this, there has been a consequential rise in the need to be able to distinguish between machine- and human-generated text across domains. Just as important is the need to be able to identify the generators for text that has been flagged as being generated by machines.

In this paper, we describe our methodology and attempts to create a system that can perform the task effectively.

\section{System Overview}
We have used RoBERTa-base for all experiments in the scope of this paper. The baseline set by the task organizers is reported to have been from a finetuned RoBERTa model. RoBERTa has the same architecture as BERT, but uses a byte-level BPE as a tokenizer and uses a different pretraining scheme and has become a SOTA model since its release \cite{liu2019roberta}.

\subsection{Weighted Layer Averaging}
The standard fine-tuning setup uses the [CLS] Representation of the last layer of RoBERTa. It has been shown that different layers of BERT-like models capture different levels of linguistic information, the lower layers capture lexical information and word order,the middle layers capture syntactic information, and the higher layers capture semantic and task specific information \cite{rogers2020primer}. We believe that using just the last layer representation may discard some of the syntactic and lexical information, which could be crucial for the task of detecting machine generated text. We use the weighted sum of all the token representations, where each layer is assigned a corresponding weight, trained along with the downstream task, similar to ElMo~\cite{peters2018deep}. Let $x_0, x_1, ... x_n$ be the input sequence. Roberta generates the following hidden states.
\[
\texttt{RoBERTa}([x_0, x_1, \ldots, x_n]) = H
\]
Where $H$ is a matrix consisting of hidden state vectors $\textbf{h}_i^j$ corresponding to the $j^\text{th}$ layer, and the $i^\text{th}$ token. $i = 0$ represents the embedding layer output.   
\[
H = 
\begin{bmatrix}
\textbf{h}^0_0 & \textbf{h}^0_1 & \ldots & \textbf{h}^0_n \\
\textbf{h}^1_0 & \textbf{h}^1_1 & \ldots & \textbf{h}^1_n \\
\vdots & \vdots & \ddots & \vdots \\
\textbf{h}^{12}_0 & \textbf{h}^{12}_1 & \ldots & \textbf{h}^{12}_n \\
\end{bmatrix}
\]

The standard fine-tuning setup uses $\textbf{h}_0^{12}$ which corresponds to the [CLS] token and passes it through another Feed Forward Network to get the output class probabilities. We propose averaging all of the layer hidden states. The input $\textbf{y}$ to the Feed Forward Network that produces the class probabilities is computed as follows.
$$
\textbf{y} = \frac{1}{12} \cdot \sum_{j=0}^{12} \frac{\lambda_j \sum_{i=0}^n \textbf{h}_i^j}{n}
$$
$\lambda_j$ is the layer weight assigned to the layer j. [$\lambda_0, \lambda_1, ... \lambda_{12}$]  are trained along with the classification task.

\subsection{Parameter Efficient Tuning with AdaLoRa}
\label{sect:pdf}

\begin{table*}
\centering
\begin{tabular}{ll}
\hline
\textbf{Model} & \textbf{Accuracy}\\
\hline
\verb|Ours| & \verb|0.7535| \\
\verb|Baseline| & \verb|0.8846| \\
\hline
\end{tabular}
\caption{\label{resultAOrg}
Results for Subtask A as computed by the organizers
}
\end{table*}

\begin{table*}
\centering
\begin{tabular}{ll}
\hline
\textbf{Model} & \textbf{Accuracy}\\
\hline
\verb|Ours| & \verb|0.7387| \\
\verb|Baseline| & \verb|0.7460| \\
\hline
\end{tabular}
\caption{\label{resultBOrg}
Results for Subtask B as computed by the organizers
}
\end{table*}

\begin{table*}
\centering
\begin{tabular}{lllll}
\hline
\textbf{Dataset} & \textbf{Precision} & \textbf{Recall} & \textbf{Accuracy} & \textbf{F1 Score}\\
\hline
\verb|Our Dev| & \verb|0.9841| & \verb|0.9949| & \verb|0.9900| & \verb|0.9895| \\
\verb|Official Dev| & \verb|0.9744| & \verb|0.9444| & \verb|0.9598| &\verb|0.9592| \\
\verb|Official Test| & \verb|0.6823| & \verb|0.9942| & \verb|0.7538| & \verb|0.8092| \\
\hline
\end{tabular}
\caption{\label{resultAOur}
Results for Subtask A as computed by us
}
\end{table*}

\begin{table*}
\centering
\begin{tabular}{lllll}
\hline
\textbf{Dataset} & \textbf{Precision}$_\text{Micro}$ & \textbf{Recall}$_\text{Micro}$ & \textbf{Accuracy} & \textbf{F1 Score}$_\text{Micro}$\\
\hline
\verb|Our Dev| & \verb|0.979| & \verb|0.979| & \verb|0.979| & \verb|0.979| \\
\verb|Official Dev| & \verb|0.9783| & \verb|0.9783| & \verb|0.9783| & \verb|0.9783| \\
\verb|Official Test| & \verb|0.7398| & \verb|0.7398| & \verb|0.7398| & \verb|0.7398| \\
\hline
\end{tabular}
\caption{\label{resultBOur}
Results for Subtask B as computed by us
}
\end{table*}

A full continual finetune of RoBERTa (and LLMs, in general) with all the weights being updated is known to potentially lead to catastrophic forgetting \cite{ramasesh2022effect}, which may cause the model to become unable to generalize, with the pretraining being, for all intents and purposes, in vain. 

It has also been shown that common pre-trained models have a very low intrinsic dimension; in other words, there exists a low dimension reparameterization that is as effective for fine-tuning as the full parameter space \cite{aghajanyan2020intrinsic}. This implies that full continual finetuning -- being potentially harmful as well as unnecessary -- can be replaced with a better, more parameter efficient method, which grants us more freedom with regards to model and data sizes. 

Low-rank Adapters (LoRA) \cite{hu2021lora} were designed with this in mind. LoRA freezes the pretrained model weights and injects trainable rank decomposition matrices into each layer of the Transformer architecture, greatly reducing the number of trainable parameters for downstream tasks. They also offer an improvement over unfreezing just the last few layers by attaching to every layer in the model, which allows them to modify information flow at every step, starting from the source.

For our task, we made use of Adaptive LoRA (AdaLoRA) \cite{zhang2023adalora}, which adjusts the matrices based on parameters learned during training, i.e. the ranks of the adapters themselves are learned. Our hope is that by doing this, we prevent unnecessarily large adapters where there is not much to do, and conversely provide the flexibility to have larger matrices to handle greater amounts of information change.

\section{Data}

Data for the task was provided by the organizers\cite{semeval2024task8}. It is an extension of the M4 Dataset \cite{wang2024m4}. The name stands for multi-generator, multi-domain, and multi-lingual corpus for machine-generated text detection. As the name suggests, the dataset has been created with text from different generators spanning multiple domains. The data for subtask A and B follow the same format, consisting of source (such as Wikipedia), model (such as Dolly), label (such as Human), and the text to be classified. The data for subtask C contains text with a combination of human- and machine-generated text, and a label indicating the word index at which the split occurs.

For our experiments, we resplit the training and dev datasets and split them uniformly across generators and domains in an 80-20 split`. Our split of the dev set is bigger than the official dev set, to get a better estimate of our model's performance.

\section{Experimental Setup}
We use RoBERTa's tokenizer and trained our models for Subtask A (monolingual) (Binary Classification) and Subtask B (Multi-Class Classification) on the resplit train data and use the resplit evaluation data for early stopping. Our Hyperparameter Configuration has been specified in Appendix \ref{sec:appendix}.

\section{Results}
Our model while doing really well on our evaluation set, falls short on the test set scoring around 13 percentage points lower than the baseline for subtask A and around 1 percentage point lower than the baseline for subtask B. This could be attributed to the model not being as good in generalising to unseen domains and generators. We hypothesize more hyperparameter tuning, better aggregation of the token representations than averaging by utilizing models like LSTMs~\cite{HochSchm97}, may help the model better generalise to unseen domains and generators by being able to capture more complex features and patterns. The submission scores as computed by the task organizers have been reported in Tables \ref{resultAOrg} and \ref{resultBOrg}. Scores on our Validation, the official validation and the official test set as computed by us have been reported in Tables \ref{resultAOur} and \ref{resultBOur}.

\section{Conclusion}

We have demonstrated that linguistic information encoded in the various layers of large language models such as RoBERTa can be used to effectively demonstrate if a text is machine-generated or not, across different domains and generators.



\bibliographystyle{acl}
\bibliography{bibliography.bib}

\appendix

\section{Hyperparameters}
\label{sec:appendix}

\begin{table*}
\centering
\begin{tabular}{lll}
\hline
\textbf{Hyperparameter} & \textbf{Value}\\
\hline
\verb|Learning Rate| & \verb|5e-4|  \\
\verb|Batch Size| & \verb|8| \\
\verb|Weight Decay| & \verb|5e-5| \\
\verb|Warmup Ratio| & \verb|0.1| \\
\verb|init_r| & \verb|12| \\
\verb|target_r| & \verb|8| \\
\verb|lora_alpha| & \verb|200| \\
\verb|lora_dropout| & \verb|0.4| \\
\hline
\end{tabular}
\caption{\label{hpa}
Hyperparameters for Subtask A (Monolingual)
}
\end{table*}

\begin{table*}
\centering
\begin{tabular}{lll}
\hline
\textbf{Hyperparameter} & \textbf{Value}\\
\hline
\verb|Learning Rate| & \verb|5e-4|  \\
\verb|Batch Size| & \verb|8| \\
\verb|Weight Decay| & \verb|5e-5| \\
\verb|Warmup Ratio| & \verb|0.01| \\
\verb|init_r| & \verb|12| \\
\verb|target_r| & \verb|8| \\
\verb|lora_alpha| & \verb|200| \\
\verb|lora_dropout| & \verb|0.4| \\
\hline
\end{tabular}
\caption{\label{hpb}
Hyperparameters for Subtask B
}
\end{table*}

\def\thefootnote{$\dagger$}\footnotetext{The code used can be found in this repository: \href{https://github.com/advin4603/AI-Detection-With-WLA}{\texttt{https://github.com/advin4603/ \\ AI-Detection-With-WLA}}} 

\end{document}